# Survey: Natural Language Parsing For Indian Languages

**Monika T. Makwana[A], Deepak C. Vegda[B]**

[A] Department of Information Technology, Dharmsinh Desai University, Nadiad, India
[B] Department of Information Technology, Dharmsinh Desai University, Nadiad, India

*Abstract*

*Syntactic parsing is a necessary task which is required for NLP applications including machine translation. It is a challenging task to develop a qualitative parser for morphological rich and agglutinative languages. Syntactic analysis is used to understand the grammatical structure of a natural language sentence. It outputs all the grammatical information of each word and its constituent. Also issues related to it help us to understand the language in a more detailed way. This literature survey is groundwork to understand the different parser development for Indian languages and various approaches that are used to develop such tools and techniques. This paper provides a survey of research papers from well known journals and conferences.*

## 1. INTRODUCTION

Syntactic analysis is the process of analyzing and determining the grammatical structure of a sentence with respect to a given formal grammar. Syntactic Parsing of a natural language sentence is considered to be an important intermediate stage for semantic analysis that can influence many pipelined application of Natural Language Processing such as information extraction, word sense disambiguation etc.

The study of structure of sentence is called syntax. It attempts to describe the grammatical order in a particular language in term of rules which in detail explain the underlying structure and a transformational process. Syntax provides rules to put together words to form components of sentences and to put together these components to form meaningful sentences. Because of the substantial ambiguity present in the human language, whose usage is to convey different semantics, it is much difficult to design the features for natural language processing tasks. The main challenge is the inherent complexity of linguistic phenomena that makes it difficult to represent the effective features for the target learning models [27].

India is a country having variety of languages major one are Indo-Aryan Languages and Dravidian Languages. Some Corpus based NLP tasks for popular languages like English, Greek etc has been worked with success. On the contrary, very little has been done on Indian languages. One of the main reasons is that not any annotated ready to use corpus sources available for such languages. Also Indian languages are morphologically rich and agglutinative in nature that makes task of creating efficient language specific tool difficult.

This paper is organized as follows; in Section 2 we present a background theory. Section 3 presents a literature review for Indian languages. In section 4 we described the measures used to evaluate a syntactic parser and at last we conclude our paper.

## 2. BACKGROUND THEORY

There exist many natural language Parsing techniques. These techniques are mainly categorized into three categories: (i) rule based (ii) statistical based and (iii) generalized parsers. All the developed parsers belong to any one of these categories and follow either 'top-down' or 'bottom-up' approach. Statistical parsing techniques are called "data-driven" and rule based parsing techniques are called "grammar-driven" approaches [27].

### A. Rule Based Parser

In rule-based approach the language specific rules are formulated to identify the best parse tree for a given grammar. But in this approach, as the production rules are applied recursively which results in overlapping. The problem can be solving efficiently by using Dynamic programming (DP) technique. The cache for sub parse trees in the DP-based parsers is called the 'chart' and consequently the DP-based parsers are called 'chart parsers'. The CYK algorithm and Early algorithm belong to rule based parsers.

*B. Statistical Based Parser*

Statistical parsing algorithms collect statistical data from correctly parsed sentences and resolves ambiguity by experience. The advantage of statistical approach is that it covers the whole grammar usage of the language. The performance of the statistical parsers depends on training corpus used to gather statistical information about the grammar of the language. Instead of using rules, statistical parsers choose the best parse tree from possible candidates based on the statistical information. The disadvantage of this approach is that sometimes comes up with invalid sequence of parse.CFG and Probabilistic Context Free Grammar (PCFG) based parsers are the examples for statistical parsers.

*C. Generalized approach*

The framework behind both rule based and statistical parsing are similar. Using this advantage, Melamed suggested another generalized parsing algorithm which was based on semi ring parsing idea. This generalized algorithm consists of five components such as: grammar, logic, semi ring, and search strategy and termination condition. In which, grammar defines terminal and non-terminal symbols, as well as a set of production rules. Logic defines the mechanism of how the parser runs by generating new partial parse trees. The semi ring defines how partial parse trees are scored. The search strategy defines the order in which partial parse trees are processed and the termination condition defines when to stop the logic necessarily.

### 3. Literature survey For Indian Languages

As compare to foreign languages, a very little work has been done in the natural language processing for Indian languages. Various Parsers for Indian Languages like Hindi, Marathi, Bengali, Kannada, Telugu and Assamese are available but it is still an ongoing process for Indian languages. One of the important measures of any parser is accuracy so that accuracy is also discussed.

Joakim Nirve in 2009 [1] presented work to optimize Malt Parser for three Indian languages Hindi, Bangla and Telgu in NLP Tools Contest at ICON 2009. They achieved second rank among participated systems. It is observed that improved labeled attachment scores are by 7-13 percent points and unlabelled attachment scores by 2-5 percent points. A small test set of 150 sentences was used to analyze the performance of the system. The performance of the system was slightly better for Bangla and Hindi languages but for Telugu it was lower than the baseline results It is observed that sustainable improvement in accuracy can be achieved by increasing the size of training data set.

Prashanth Mannem in 2009[7] proposed a bidirectional dependency parser for Hindi, Telugu and Bangla languages The developed parser uses a bidirectional parsing algorithm with two operations projection and non-projection to build the dependency tree. The performance of the parser was evaluated based on the test data sentences. He reported that the system achieves a labeled attachment score of 71.63%, 59.86% and 67.74% for Hindi, Telugu and Bangla respectively on the Treebank with fine-grained dependency labels. Based on the coarse-grained labels the dependency parser achieved 76.90%, 70.34% and 65.01% accuracies respectively.

Bharat Ram Ambati et al. in 2009 [17] explored two data-driven parsers called Malt and MST on three Indian languages namely Hindi, Bangla and Telugu. They merged both the training and development data and did 5-fold cross-validation best settings from the cross validation experiments and these settings are applied on the test data of the contest. Finally they evaluated the individual and average results on both coarse-grained and fine-grained tag set for these three languages. They found that for all the languages Malt performed better over MST+maxent. They also modified the implementation of MST to handle vibhakti and TAM markers for labeling. They reported that, the average of best unlabeled attachment, labeled attachment and labeled accuracies are 88.43%, 71.71% and 73.81% respectively

Akshar Bharati et al. in 2009 [2] proposed a simple parser for Indian languages in a dependency framework. They describe syntactic parser which follows a grammar driven methodology. They described a grammar oriented model that makes use of linguistic features to identify relations. The proposed parser was modeled based on Paninian grammatical approach which provides a dependency grammar framework. They also compared the proposed parser performance against the previous similar attempts and reported its efficiency. They had compared its performance against previous similar attempts and reported its efficiency. They have showed how by using simple yet robust rules one can achieve high performance in the identification of various levels of dependency relations.

Meher Vijay Yeleti and Kalyan Deepak in 2009 [6] developed a constraint based Hindi dependency parser. In the proposed system a grammar driven approach was complemented by a controlled statistical strategy to achieve high performance and robustness. The developed system uses two stage constraint based hybrid approach to dependency parsing. They have defined two stages and this division leads to selective identification and resolution of specific dependency relations at the two stages. They also used hard constraints and soft constraints to build an efficient and robust hybrid parser. From the experiment they found out that the best labeled and unlabeled attachment accuracies for Hindi are 62.20% and 85.55% respectively.

Phani Gadde et al. in 2010[9] describes a data driven dependency parsing approach which uses information about the clauses in a sentence to improve the performance of a parser. The clausal information is added automatically using a partial parser. They demonstrated the experiments on Hindi, a morphologically rich, free-word-order language, using a modified version of MST Parser. They did all the experiments on the ICON 2009 parsing contest data. They achieved an improvement of 0.87% and 0.77% in unlabeled attachment and labeled attachment accuracies respectively over the baseline parsing accuracies.

Bharat Ram Ambati et al. in 2010 [8] analyzes the relative importance of different linguistic features for data-driven dependency parsing of Hind**i**. The analysis shows that the greatest gain in accuracy comes from the addition of morph syntactic features related to case, tense, aspect and modality. They had combined the features from the two parsers and achieved a labeled attachment score of 76.5%, which is 2 percentage points better than the previous state of the art.

Swati Ramteke et al in 2014[13] developed lexicon Parser for Devanagari script (Hindi), it shows how a Hindi sentence is parsed into tokens and then find the relationship between tokens using grammar and by using semantic representation generate a parse tree. They used Rule based approach to resolves the disambiguity of words. Tagging and tokenization algorithms were developed and implemented for Devanagari text. The accuracy of 89.33% was achieved from Lexicon parser. From the experiments, it has been observed that the accuracy was low when they tested more ambiguity sentences and sentences of future tense. Similarly, when they tested sentences of simple present and past tenses then the accuracy was very high.

Sankar De et al. in 2009 [16] proposed a constraint based Dependency parsing system and applied to a free-word order language Bangla . They have used a structure simplification and demand satisfaction approach to dependency parsing in Bangla language. A well known and very effective grammar formalism for free word order language called Paninian Grammatical model was used for this purpose. The basic idea behind this approach is to simplify the complex and compound sentential structures first, then to parse the simple structures so obtained by satisfying the 'Karaka' demands of the Demand Groups (Verb Groups) and to rejoin such parsed structures with appropriate links and Karaka labels. A Treebank of 1000 annotated sentences was used for training the system. The performance of the system was evaluated with 150 sentences and accuracies achieved are of 79.81%,90.32%, 81.27% and   for labeled attachments, unlabeled attachments and label scores respectively.

Aniruddha Ghosh et al. in 2009 [4] proposed a dependency parser system for Bengali language. They have performed two separate experiments for Bengali. Statistical CRF based model followed by a rule-based post-processing technique has been used. They have used ICON 2009 datasets for training the system. The probabilistic sequence model trained with the morphological features like root word, , chunk tag, vibhakti, POS-tag and dependency relation from the training set data. The result of the baseline CRF based system is filtered by a rule-based post-processing module by using the output obtained through the rule based dependency parser. The system demonstrated an unlabeled attachment score (UAS) of 74.09%, labeled attachment score (LAS) of 53.90% and labeled accuracy score (LS) of 61.71% respectively.

Sanjay Chatterji and et al. in 2009 [5] proposed a hybrid approach for parsing Bengali sentences. The proposed system was based on data driven dependency parser. In order to improve the performance of the system, some hand-crafted rules are identified based on the error patterns on the output of the baseline system.

Akshar Bharati and Rajeev Sangal described a grammar formalism called the 'Paninian Grammar Framework' that has been successfully applied to all free word Indian languages [14]. They have described a constraint based parser. Paninian framework uses the notation of karaka relations between verbs and nouns in a sentence. It is found that the Paninian framework applied to modern Indian languages will give an elegant account of the relation between vibhakti and karaka roles and that the mapping is elegant and compact

In the year 2009, B.M. Sagar et al proposed a way of producing context free grammar for the Noun Phrase and Verb Phrase agreement in Kannada Sentences [11]. In this approach, a recursive descent parser is used to parse the context free grammar. The system works in two levels: First of all, it generates the CFG of the sentence. In the second level, a recursive descent parser called Recursive Descent Parser of Natural Language Tool Kit (NLTK) was used to test the grammar. In short, it is a grammar checking system in which for a given sentence parser determines whether the sentence is syntactically correct or wrong depending upon the Noun and Verb agreement. They have tested the system with 200 sample sentences and obtained encouraging results.

Antony P J et al. in 2010 [10] have developed a Penn Treebank based statistical syntactic parsers for Kannada language. The well known grammar formalism called Penn Treebank structure was used to create the corpus for proposed statistical syntactic parser. The parsing system was trained with 1,000 Kannada sentences. The developed corpus has been already annotated with correct segmentation and Part-Of-Speech information. The developers used their own SVM based POS tagger generator for assigning proper tags to each and every word in the training and testing sentences. The proposed syntactic analyzer was implemented using supervised machine learning and probabilistic context free grammars approaches. Training, testing and evaluation were done by support vector method (SVM) algorithms. Experimental observations show that the performance of the proposed system is significantly good and has very competitive accuracy.

B.M. Sagar et al. in 2010[15] proposed a Context Free Grammar (CFG) analysis for simple Kannada sentences. They have explained the writing of Context Free Grammar (CFG) for a simple Kannada sentence with two types of examples. In the developed system, a language grammar is parsed with Top-Down and Bottom-Up parsers and they found that a Top-Down parser is more suitable to parse the given grammatical production.

Rahman, Mirzanur and et al. in 2009 [3] have developed a context free grammar for simple Assamese sentences. In these work they had considered only limited number of sentences for developing rules and only seven main tags are used. They have analyzed the issues that arise in parsing Assamese sentences and produce an algorithm to solve those issues. The algorithm is a modification of Earley's Algorithm and they found the algorithm simple and efficient.

Navanath Saharia et al. in 2011 [12] described a parsing criterion for Assamese text. They have described the practical analysis of Assamese sentences from a computational perspective. This approach can be used to parse the simple sentences with multiple noun, adjective, adverb clauses.

Dhanashree Kulkarni et al. in 2014[18] has made an attempt to write context free grammar for simple Marathi sentences. Two sets of examples are taken to explain the writing of CFG. Grammar is parsed with Top Down and Bottom-Up Parser. Top Down parser is said to be more suitable to parse grammatical productions This paper sets a stage to develop computerized grammar checking methods for a given Marathi sentence and stresses mainly on representation of CFG considered.

B. Venkata S. kumari et al in 2012[19] presents this paper, they first explored Malt and MST parsers and developed best models, which they considered as the baseline models for their approach. Considering pros of both these parsers, they developed a hybrid approach combining the output of these two parsers in an intuitive manner. They showed that a simple system like combining both MST and Malt in an intuitive way can perform better than both the parsers. They reported their results on both development and test data provided in the Hindi Shared Task on Parsing at workshop on MT and parsing in Indian Languages, Coling 2012. Their system's secured labeled attachment score of 90.66% and 80.77% for gold standard and automatic tracks respectively. The accuracies are 3rd best and 5th best for gold standard and automatic tracks respectively.

In this paper, Sambhav jain et al. in 2013[20] presents an efforts towards incorporating external knowledge from Hindi Word Net to aid dependency parsing. They conduct parsing experiments on Hindi, utilizing the information from concept ontologies available in Hindi Word Net to complement the morph syntactic information already available. The work is driven by the insight that concept ontologies capture a specific real world aspect of lexical items, which is quite distinct and unlikely to be deduced from morph syntactic information such as morph, POS-tag and chunk. This complementing information is encoded as an additional feature for data driven parsing and experiments are conducted. They perform experiments over datasets of different sizes. They achieve an improvement of 1.1% (LAS) when training 1,000 sentences and 0.2% (LAS) on 13,371 sentences over the baseline. The improvements are statistically significant at $p<0.01$. The higher improvements on 1,000 sentences suggest that the semantic information could address the data sparsity problem.

Pradipta Ranjan et al. in 2003[21] presented an algorithm for local word grouping to extricate fixed word order dependencies in Hindi sentences. Local word grouping is achieved by defining regular expressions for the word groups. Computational Paninian model. Also, local word grouping achieved can be used to provide inputs to intonation and ambiguities occurring during word grouping are also resolved. Hindi being a free order language, fixed order word group extraction is essential for decreasing the load on the free word order parser. The parser paradigm being used is the prosody modeling units for text to speech systems in Indian languages. Part of speech tagging is an essential requirement for local word grouping. They present another algorithm for part of speech tagging based on lexical sequence constraints in Hindi. The algorithm acts as the first level of part of speech tagger, using constraint propagation, based on ontological information and information from morphological analysis, and lexical rules.

In this paper karan singla et al. [22] has experimented with different parameters of data-driven Malt Parser along with the two-stage preprocessing approach to build a high quality dependency parser for Hindi. The system acheived best LAS of 90.99% for gold standard track and second best LAS of 83.91% for automated data.

Selvam M et al. in 2008 [23] proposed a statistical parsing of Tamil sentences using phrase structure hybrid language model. They have built a statistical language model based on Trigram for Tamil language with medium of 5000 words. In the experiment they showed that statistical parsing gives better performance through trigram probabilities and large vocabulary size. In order to overcome some disadvantages like focus on semantics rather than syntax, lack of support in free ordering of words and long term relationship of the system, a structural component is to be incorporated. The developed hybrid language model is based on a part of speech tag set for Tamil language with more than 500 tags. The developed phrase structured Treebank was based on 326 Tamil sentences which covers more than 5000 words. The phrase structured Treebank was trained using immediate head parsing technique. Two test cases with 120 and 40 sentences have been selected from trained set and test set respectively. They reported that, the performance of the system is better than the grammar model.

Bharati, Akshar, et al. in 2008[24] presented a paper is an attempt at exploring and isolating some crucial cues present in the language which lend themselves to robust dependency parsing. They report a series of experiments. In the process of these experiments they also compared the performance of two freely available dependency parsers and pointed their strengths and weaknesses. The results obtained validate various linguistic intuitions which can be effectively used in parsing. In particular we note that conjoined vibhakti-label feature and minimal semantics can lead to drastic improvement in the parser performance. Apart from this the results also point towards some hard to learn linguistic constructions.

**Table 1 Literature survey**

| Sr no. | Papers name (year) | Publication details And (Author names) | Language | Method/ Algorithm/ Tool | Accuracy | Corpus/ Dataset |
|---|---|---|---|---|---|---|
| 1 | Parsing indian languages with maltparser (2009) | Proceedings of the ICON09 NLP Tools Contest: Indian Language Dependency Parsing : 12-18. (Joakim Nirve) | Hindi, Bagla, Telgu | Transition-based approach MALT parser | UAS:H-90%, B-90% and T-85% LAS:15-25% low | Training set of: Hindi-1651 Bangla-1130 Telgu-1615 Test sentences-150 |
| 2 | Simple parser for Indian languages in a dependency framework. (2009) | Proceedings of the Third Linguistic Annotation Workshop. Association for Computational Linguistics. (Akshar Bharati et al.) | Hindi | Grammar driven approach | Precision-96.2% Recall -82.6% | Hyderabad dependency treebank Total 2100 words Trainingset 1300 Testset 800 |
| 3. | Parsing of part-of-speech tagged Assamese Texts (2009) | IJCSI International Journal of Computer Science Issues, Vol. 6, No. 1, 2009 (Rahman, Mirzanur et al.) | Assamese | Earley's Algorithm | Earley's algorithm is simple and effective | Assamese sentences |
| 4 | Dependency Parser for Bengali: the JU System at ICON 2009 (2009) | Proceedings of ICON09 NLP Tools Contest: Indian Lan-guage Dependency Parsing, Hyderabad, India, 2009. (Aniruddha Ghosh et al.) | Bengali | Rule Based | UAS-74.09% LAS-53.90% LS-61.71% | ICON 2009 datasets |
| 5 | Grammar Driven Rules for Hybrid Bengali Dependency Parsing (2009) | Proceedings of ICON-2009 7th International Conference on Natural Language Processing, Macmillan Publishers, India (Sanjay Chatterji et al.) | Bengali | Hybrid approach MALT parser | Highly effective rules | ICON 2009 datasets |
| 6 | Constraint based Hindi dependency parsing (2009) | Proceedings of ICON09 NLP Tools Contest: Indian Lan-guage Dependency Parsing, Hyderabad, India, 2009. (Meher Vijay Yeleti, Kalyan Deepak) | Hindi | Hybrid approach | LSA-62.20 UAS -85.55 | ICON 2009 datasets hindi data |

Monika et al                                                                 Survey: Natural Language Parsing For Indian Languages

| | | | | | | |
|---|---|---|---|---|---|---|
| 7 | Bidirectional Dependency Parser for Hindi, Telugu and Bangla (2009) | Proceedings of ICON09 NLP Tools Contest: Indian Language Dependency Parsing, Hyderabad, India, 2009. (Prashanth Mannem) | Hindi, Telugu Bagla, | Bidirection Dependency parser algo | LAS for: Hindi-71.63% Telgu-59.86 % bangla- 67.74% | ICON 2009 datasets |
| 8 | On the role of morphosyntactic features in Hindi dependency parsing (2010) | Proceedings of the NAACL HLT 2010 First Workshop on Statistical Parsing of Morphologically-Rich Languages. Association for Computational Linguistics, 2010 (Bharat Ram Ambati et al.) | Hindi | Data Driven | LAS-76.5% | ICON 2009 datasets |
| 9 | Improving data driven dependency parsing using clausal information (2010) | 11th Annual Conference of the North American Chapter of the Association for Computational Linguistics (NAACL-HLT, 2010) (Phani Gadde et al.) | Hindi | Data Driven dependency approach | LAS-74.39% UAS- 91.87% LS- 76.21% | ICON 2009 contest data |
| 10 | Penn Treebank-Based Syntactic Parsers for South Dravidian Languages using a Machine Learning Approach (2010) | International journal on Computer Ap-plication (IJCA), No. 08, ISBN: 978-93-80746-92-0, 2010. (Antony P J) | Kannad | Stastical parser | Good | Penn treebank 1000 kannad sentences |
| 11 | Solving the Noun Phrase and Verb Phrase Agreement in Kannada Sentences (2009) | In-ternational Journal of Computer Theory and Engineering, Vol. 1, No. 3, August, 2009, 1793-8201 (B.M. Sagar et al.) | Kannad | Recursive Descent Parser | Good | 200 sentences |
| 12 | A First Step Towards Parsing of Assamese Text (2011) | Special Volume: Problems of Parsing in Indian Languages (Navanath Saharia et al.) | Assamese | Rule Based | 78.82% | ICON 2009 datasets |
| 13 | Lexicon Parser for syntactic and semantic analysis of Devanagari sentence using Hindi wordnet (2014) | International Journal of Advanced Research in Computer and Communication Engineering Vol. 3, Issue 4, April 2014 (Swati Ramteke et al.) | Hindi | Rule based | 89.33% | 1500 tokens |
| 14 | Parsing Free Word Order Languages in the Paninian Framework (2009) | (Akshar Bharati, Rajeev Sangal) | hindi | Constraint based parser | Efficient and effective parser | ICON 2009 datasets |
| 15 | Context Free Grammar (CFG) Analysis for simple Kannada sentences (2010) | Special Issue of IJCCT Vol.1 Issue 2, 3, 4; 2010 for International Conference [ACCTA-2010], 3-5 August 2010.. (B.M. Sagar et al.) | Kannada | Both TopDown and Bottom-Up parsers | Top-Down parser is more efficient | Kannad sentences |
| 16 | Structure Simplification and Demand Satisfaction Approach to De-pendency Parsing in Bangla (2009) | Proceedings of ICON09 NLP Tools Contest:Indian Language Dependency Parsing, Hyderabad, India (Sankar et al.) | Bangla | Constrained based dependency parser | LAS-79.81% UAS- 90.32% LS- 81.27% | Treebank dataset training sentences- 1000 test sentences- 150 |
| 17 | Experiments in Indian Language Dependency Parsing (2009) | Proceedings of ICON09 NLP Tools Contest: Indian Language Dependency Parsing, Hyderabad, India, (Bharat Ram Ambati et al.) | Hindi, Bangla Telugu | Data driven-MST and MALT parsers | UAS-88.43% LAS-71.71% LS- 73.81% | ICON 2009 datasets |
| 18 | Specifying context free grammar for marathi sentences (2014) | International Journal of Computer Applications 99.14 (2014): 38-41 (Dhanashree Kulkarni et al.) | Marathi | Both top-down and bottom-up parsers | Top-down parser is mare efficient | Simple Marathi sentences |
| 19 | Hindi dependency parsing using a combined model of MALT and MST (2012) | Proceedings of the Workshop on Machine Translation and Parsing in Indian Languages (MTPIL-2012), pages 171–178, COLING 2012, Mumbai, December 2012. (B. Venkata S. kumari et al.) | Hindi | Hybrid approach(MALT +MST) | (for gold standard) LAS-90.66% (for automatic tracks) LAS- 80.77% | gold standard automatic tracks |
| 20 | Exploring Semantic Information in Hindi WordNet for Hindi Dependency Parsing (2013) | The sixth international joint conference on natural language processing (IJCNLP2013). (Sambhav jain et al.) | Hindi | HWN ontology | Better accuracy is achieved | Hindi Dependency Treebank |
| 21 | Part of speech tagging and local word grouping techniques for natural language parsing in Hindi (2003) | Proceedings of the 1st International Conference on Natural Language Processing (ICON 2003) (Pradipta Ranjan et al.) | Hindi | Paninian model | Improved performance of a parser | Hindi sentences |
| 22 | Two-stage approach for hindi dependency parsing using maltparser (2012) | Proceedings of the Workshop on Machine Translation and Parsing in Indian Languages (MTPIL-2012), pages 163–170, COLING 2012, Mumbai, December 2012. (Karan singla et al.) | Hindi | MALT parser | gold standard track LAS-90.99% Automated data LAS-83.91% | gold standard track Automated data |
| 23 | Hindi Parser- Based on CKY algorithm | Int. J. Computer Technology and Applications, Vol. 3(2), 851-853 (Nitin Hambir and Ambrish Shrivastav) | Hindi | CKY algorithm | Good | Hindi Sentences |
| 24 | Structural Parsing of Natural Language Text in Tamil Using Phrase Structure Hybrid Language Model (2008) | International Journal of Computer, Information and Systems Science, and Engineering (2008): 2-4. (Selvam M et al.) | Tamil | phrase structure hybrid language model | Good | 5000 words |
| 25 | Two semantic features make all the difference in parsing accuracy (2008) | In Proceedings of the 6th International Conference on Natural Language Processing (ICON-08), CDAC Pune, India. 2008. (Bharati, Akshar, et al.) | Hindi | MALT and MST parser | LAS-69.64% UAS-88.67% | Hindi Dependency treebank 1200 sentences |
| 26 | Two methods to incorporate local morphosyntactic features in Hindi dependency parsing (2010) | Proceedings of the NAACL HLT 2010 First Workshop on Statistical Parsing of Morphologically-Rich Languages. Association for Computational Linguistics, 2010. (Bharat Ram Ambati et al.) | Hindi | MALT and MST parser | Good | Hindi Treebank |

## 4. Issues in Syntactic Parsing

Indian languages are Resource poor and less privileged where annotated corpora is not available, so to create tagged corpus is tedious and time consuming task. Hence developing a well syntactic parser is a challenging task.

The challenges in syntactic analysis of a text are called **structural ambiguity.** Different types of challenges during syntactic analysis are as follow:

i. <u>Scope Ambiguity</u> The first level of ambiguity is scope ambiguity.

**Ram is eating food and watching television**.

The scope of the subject "Ram" is ambiguous. The question in the sentence is "to which activity Ram refers to i.e. whether Ram is only eating food or he is watching television or doing both the activities. How much of the text subject is qualified? This is known as scope ambiguity means what is the region of influence i.e. the scope of the subject here. How much text does it qualifies?

ii. <u>Attachment ambiguity</u> Attachment ambiguity arises from uncertainty of attaching a phrase or clause to a part of a sentence. Here are some examples:

**I saw the girl with a telescope.**

It is not clear who has the telescope, I or the Girl? In the former case, we say, the preposition phrase "with a telescope" attaches with the verb "saw" with the instrumental case. In the latter the PP attaches to "the girl" as a modifier.

Indian languages have post positions instead of prepositions, that is, entities that assign case roles follow the noun, and do not precede. In case of Hindi:

### दूरबीन से लड़की को देखा

Durbin se ladki ko dekha

Telescope with girl_ ACC saw

saw the girl with a telescope

The postposition "se" assigns case role to "Durbin" and follows it.

Attachment ambiguity of the type pp-attachment is not so common in Indian languages which are as a rule SOV (subject-object-verb) languages. Postpositions follow this pattern:

NP1 P NP2 V

(In above example, NP1= Durbin, P=se, NP2= ladki, V= dekha)

Postpositions typically assign case and hardly modify the following NP. One exception to this is the genitive case (of; Hindi का के की ka, ke, kii). But the genitive case marker always links two NPs.

### पूजा ने पूजा के लिए फूल तोड़ा

Pooja ne pooja ke liye phool toda

pooja_ERG worship_for flowers plucked

Pooja plucked flowers for worship

Here the first pooja is the name of a girl and the second pooja means worship. Translating this as worship plucked flowers for worship is not correct, though pooja plucked flowers for pooja is passable [26].

## 5. Performance measures

The Performance of any Syntactic Parser is evaluated by using the measures such as Precision, Recall, and F-measure etc. In this regards, different number of sentences are used for testing and the training corpus are gather from Word Net.

During training and testing data, the parser was trained on the entire released data with the best performing feature set and the un-annotated test data was parsed with the model obtained.

### Conclusion

In this paper work, we have presented a survey on development of different Syntactic parsers for Indian languages with their performance. Also we tried to discuss in brief some existing approaches that have been used to develop parsers for Indian languages. It shows the clear need of annotated corpus for different Indian languages to develop efficient Syntactic Analyzers. Although now some Indian languages had prepared annotated corpora still rest of are still suffering from the problem of lacking annotated corpus. Our Future work is to create annotated corpora as well as an efficient Syntactic Analyzer by considering the agglutinative and morphological rich features of language to donate our some sort of contribution to resource poor Indian language.

## References

[1] Nivre Joakim.(2009), Parsing indian languages with maltparser, *Proceedings of the ICON09 NLP Tools Contest*: Indian Language Dependency Parsing : 12-18.

[2] Bharati, Akshar, et al.(2009), Simple parser for Indian languages in a dependency framework, *Proceedings of the Third Linguistic Annotation Workshop.* Association for Computational Linguistics.

[3] Rahman, Mirzanur Sufal Das and Utpal Sharma.(2009) Parsing of part-of-speech tagged Assamese Texts, *IJCSI International Journal of Computer Science Issues*, Vol. 6, No. 1,

[4] Ghosh Aniruddha, et al.(2009), Dependency Parser for Bengali: the JU System at ICON 2009, *NLP Tool Contest ICON*.

[5] S.Chatterji, P.Sonare, S.Sarkar and D.Roy.(2009),Grammar Driven Rules for Hybrid Bengali Dependency Parsing ,

*Proceedings of ICON09 NLP Tools Contest*: Indian Language De-pendency Parsing, Hyderabad, India, 2009

[6] M.Vijay Yeleti and K.Deepak.(2009),Constraint based Hindi dependency parsing, Language Technologies Research Centre*, Pro-ceedings of ICON09 NLP Tools Contest*: Indian Language Depen-dency Parsing, Hyderabad, India, 2009.

[7] Prashanth Mannem.(2009),Bidirectional Dependency Parser for Hindi, Telugu and Bangla, Language Technologies Research Center, Inter-national Institute of Information Technology, *Proceedings of ICON09 NLP Tools Contest*: Indian Language Dependency Parsing, Hyderabad, India, 2009.

[8] Ambati, Bharat Ram, et al.(2010). On the role of morphosyntactic features in Hindi dependency parsing. *Proceedings of the NAACL HLT 2010 First Workshop on Statistical Parsing of Morphologically-Rich Languages. Association for Computational Linguistics*, 2010.

[9] Gadde, Phani, et al.(2010). Improving data driven dependency parsing using clausal information. Human Language Technologies: *The 2010 Annual Conference of the North American Chapter of the Association for Computational Linguistics. Association for Computational Linguistics*, 2010.

[10] Antony P J, Nandini. J. Warrier and Soman K P(2010), Penn Treebank-Based Syntactic Parsers for South Dravidian Languages using a Machine Learning Approach*", International journal on Computer Ap-plication* (IJCA), No. 08, ISBN: 978-93-80746-92-0, 2010.

[11] B.M. Sagar, Shobha G and Ramakanth Kumar P.(2009), Solving the Noun Phrase and Verb Phrase Agreement in Kannada Sentences, *International Journal of Computer Theory and Engineering*, Vol. 1, No. 3, August, 2009, 1793-8201

[12] N.Saharia, U.Sharma and J.Kalita(2011) A First Step Towards Parsing of Assamese Text, *Special Volume: Problems of Parsing in Indian Languages* .

[13] S.Ramteke, K.Ramteke, Rajesh Dongare(2014), Lexicon Parser for syntactic and semantic analysis of Devanagari sentence using Hindi wordnet, *International Journal of Advanced Research in Computer and Communication Engineering* Vol. 3, Issue 4, April 2014

[14] Akshar Bharati and Rajeev Sangal(2009), Parsing Free Word Order Languages in the Paninian Framework, *proceeding ACL '93 Proceeding of the 31st annual meeting on Association for Computational Linguistic*

[15] B.M. Sagar, Shobha G and Ramakanth Kumar P.(2010), Context Free Grammar (CFG) Analysis for simple Kannada sentences, *Special Issue of IJCCT* Vol.1 Issue 2, 3, 4; 2010 *for International Confe-rence* [ACCTA-2010], 3-5 August 2010.

[16] Sankar, Arnab Dhar and Utpal G.(2009), Structure Simplification and Demand Satisfaction Approach to Dependency Parsing in Bangla, *Proceedings of ICON09 NLP Tools Contest*: Indian Language Dependency Parsing, Hyderabad, India, 2009.

[17] Bharat Ram Ambati, Phani Gadde and Karan Jindal(2009), Experiments in Indian Language Dependency Parsing, *Proceedings of ICON09 NLP Tools Contest*: Indian Language Dependency Parsing, Hydera-bad, India, 2009

[18] Kulkarni, Dhanashree.(2014) Specifying Context free Grammar for Marathi Sentences. *International Journal of Computer Applications* 99.14 (2014): 38-41.

[19] Kumari, B. Venkata Seshu, and Ramisetty Rajeswara Rao.(2012) Hindi Dependency Parsing using a combined model of Malt and MST. *24th International Conference on Computational Linguistics.*

[20] Jain, Sambhav, et al.(2013). Exploring Semantic Information in Hindi WordNet for Hindi Dependency Parsing. *The sixth international joint conference on natural language processing(IJCNLP2013)*.

[21] Ray, Pradipta Ranjan, et al.(2003).Part of speech tagging and local word grouping techniques for natural language parsing in Hindi. *Proceedings of the 1st International Conference on Natural Language Processing* (ICON 2003).

[22] Singla, Karan, et al.(2012). Two-stage approach for hindi dependency parsing using malt parser. *Proceedings of the Workshop on Machine Translation and Parsing in Indian Languages* (MTPIL-2012), pages 163–170,COLING 2012, Mumbai, December 2012.

[23] Selvam M., A. M. Natarajan and R. Thangarajan.(2008) Structural Parsing of Natural Language Text in Tamil Using Phrase Structure Hybrid Language Model. *International Journal of Computer, Information and Systems Science, and Engineering* (2008): 2-4.

[24] Bharati, Akshar, et al.(2008). Two semantic features make all the difference in parsing accuracy. *In Proceedings of the 6th International Conference on Natural Language Processing* (ICON-08), CDAC Pune, India. 2008.

[25] Ambati, Bharat Ram, et al.(2010). Two methods to incorporate local morphosyntactic features in Hindi dependency parsing. *Proceedings of the NAACL HLT 2010 First Workshop on Statistical Parsing of Morphologically-Rich Languages. Association for Computational Linguistics*.

[26] Pushpak Bhattacharya.(2012), Natural language Processing: A Perspective from Computation in Presence of Ambiguity, Resource Constraint and Multilinguality,*CSI journal of computing,* VOL.1 No.2,2012

[27] Antony, P. J., and K. P. Soman.(2012). Computational morphology and natural language parsing for Indian languages: a literature survey, *Int J Comput Sci Eng Technol* 3.4 (2012): 136-146